\documentclass[preprint,12pt]{elsarticle}

\usepackage{lineno}

\usepackage[inline]{enumitem}

\usepackage{booktabs}
\usepackage{subcaption}
\usepackage{multicol, multirow}

\usepackage{amssymb}
\usepackage{amsmath}

\usepackage{graphicx}

\usepackage[hidelinks]{hyperref}
\usepackage[nameinlink]{cleveref}

\usepackage{todonotes}

\biboptions{authoryear}

\journal{Expert Systems with Applications}

\begin{document}

\begin{frontmatter}

\title{\vspace{-8mm}Short-term bus travel time prediction for transfer synchronization with intelligent uncertainty handling\vspace{-2mm}}

\author[movia,dtu]{Niklas Christoffer Petersen\corref{cor1}}
\ead{niklch@dtu.dk}
\author[dtu2]{Anders Parslov}
\ead{anders@parslov.com}
\author[dtu]{Filipe Rodrigues}
\ead{rodr@dtu.dk}

\address[movia]{Movia Public Transport, 
    Gammel Køge Landevej 3,
    Valby, Denmark}

\address[dtu]{Department of Technology, Management and Economics,
   Technical University of Denmark,
   Kgs. Lyngby, Denmark
}
   
\address[dtu2]{Department of Applied Mathematics and Computer Science,
   Technical University of Denmark,
   Kgs. Lyngby, Denmark
   \vspace{-6mm}}

\cortext[cor1]{\textit{Corresponding author}}

\begin{abstract}
\footnotesize
This paper presents two novel approaches for uncertainty estimation adapted and extended for the multi-link bus travel time problem. The uncertainty is modeled directly as part of recurrent artificial neural networks, but using two fundamentally different approaches: one based on \textit{Deep Quantile Regression} (DQR) and the other on \textit{Bayesian Recurrent Neural Networks} (BRNN). Both models predict multiple time steps into the future, but handle the time-dependent uncertainty estimation differently. We present a sampling technique in order to aggregate quantile estimates for link level travel time to yield the multi-link travel time distribution needed for a vehicle to travel from its current position to a specific downstream stop point or transfer site. 

To motivate the relevance of \textit{uncertainty-aware} models in the domain, we focus on the connection assurance application as a case study: An expert system to determine whether a bus driver should hold and wait for a connecting service, or break the connection and reduce its own delay. Our results show that the DQR-model performs overall best for the 80\%, 90\% and 95\% prediction intervals, both for a 15 minute time horizon into the future ($t + 1$), but also for the 30 and 45 minutes time horizon ($t + 2$ and $t + 3$), with a constant, but very small underestimation of the uncertainty interval (1-4 pp.). However, we also show, that the BRNN model still can outperform the DQR for specific cases. Lastly, we demonstrate how a simple decision support system can take advantage of our \textit{uncertainty-aware} travel time models to prioritize the difference in travel time uncertainty for bus holding at strategic points, thus reducing the introduced delay for the connection assurance application.
\end{abstract}


\end{frontmatter}


\section{Introduction}
\label{sec:intro}

Compared to other public transport modes, bus travel time is arguably more prone to variability due to the mixture with other traffic modes (e.g.\ cars, cyclists) in a shared road infrastructure. Modeling and quantifying the uncertainty of travel time predictions has several applications, which directly or indirectly support more advanced public transport information systems, such as:
\begin{itemize}
    \item Connection assurance: The process of synchronizing two public transport services (e.g. buses, trams, trains) at the operation level, for ensuring or increasing the probability of passenger exchange between the services. Several studies have shown improvements using traditional point-estimated travel and arrival times based on real-time vehicle location data. \citep{Wu_ScheduleCoordination, Ting2007}. 
    \item Travel planner apps and traffic information systems: Allowing the system to warn the user that they may not be able to complete the entire journey in cases where a suggested travel plan consists of multiple public transport modes or has connections.
\end{itemize}

\begin{figure}[t]
    \centering
    \includegraphics[width=\textwidth]{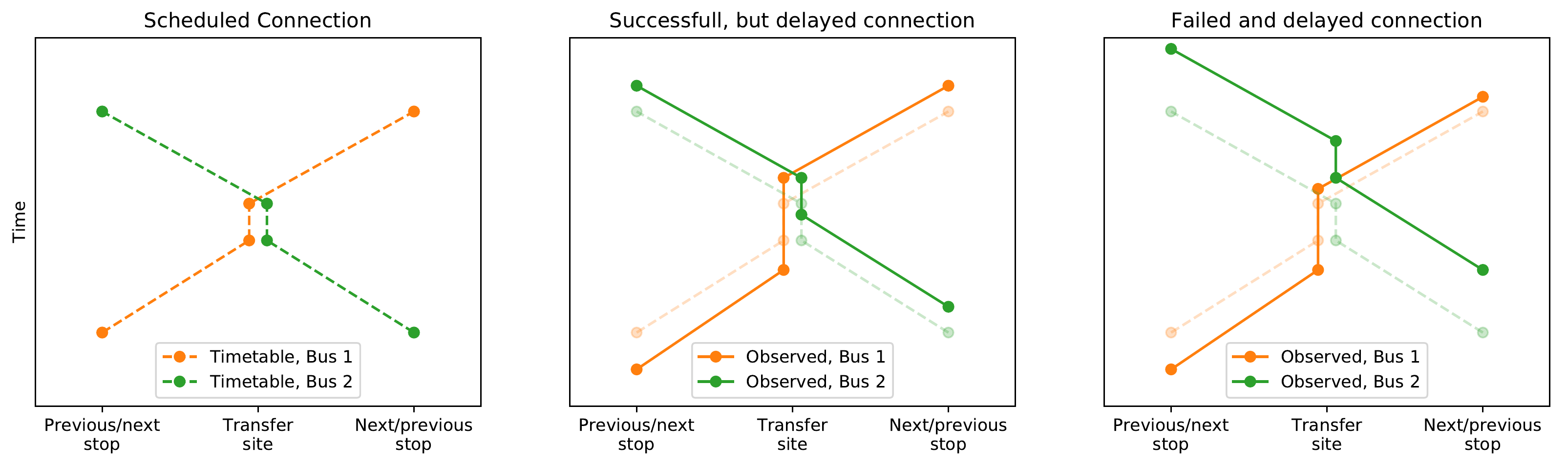}
    \caption{Examples of scheduled and observed outcomes of the bus holding strategy.}
    \label{fig:connection_assurance}
\end{figure}

To motivate and support our arguments for modelling travel time uncertainty, we focus on the connection assurance application as a case study: on the scheduling level, connection assurance is often simply modeled into the timetable as overlapping dwell time at nearby stops/stations (i.e.\ the transfer site) as illustrated by~\Cref{fig:connection_assurance} (left). Thus, there is a scheduled time window for passenger exchange between both public transport services~\citep{Ibarra_BusSynchronization, Wu_ScheduleCoordination}. The window needs to cover both passenger walking time between the stops/stations, as well as boarding and alighting time.

Delays and, in general, deviations from the schedule are a natural aspect of operating a public transport service, especially for buses and trams which often share road infrastructure with other means of transport. Thus, a frequent applied strategy is the \textit{bus holding strategy}. \Cref{fig:connection_assurance} (center) shows an example of this: \textit{Bus 1} is ahead of schedule, arrives early to the transfer site, and \textit{holds} for \textit{Bus 2}. However, \textit{Bus 2} has been delayed at previous stops and arrives at the transfer site delayed, but within the scheduled window. The passenger exchange occurs, but the result is that both \textit{Bus 1} and \textit{Bus 2} depart delayed from the transfer site. Therefore, it is evident that the decision on how long to hold and how to prioritize holding time at different strategic points have impacts on the adherence to the schedule, and thus the probability for both transferring and non-transferring passengers to reach other connections and their destination on time. In~\Cref{fig:connection_assurance} (right) \textit{Bus 1}, in the absence of \textit{Bus 2} arriving at the transfer site, decides to continue, thus breaking the connection assurance, but on the other hand ends up being less delayed than in the previous case. 

At the operation level, there are several options to implement the bus holding strategy: 
\begin{enumerate}
    \item If there is visual contact between the stops/stations at the transfer site, one naive option is to leave the enforcement of the connection assurance to the drivers. However, without any real-time knowledge of location of the opposite public transport service there are risks: e.g.\ in~\Cref{fig:connection_assurance} (right) \textit{Bus 2} might still hold for the scheduled window, because the driver has no knowledge of whether \textit{Bus 1} is also delayed, or whether \textit{Bus 1} has already departed, thus breaking the connection assurance. In the latter case, it increases the delay of \textit{Bus 2} without any proper reason.
    \item An extension to (1) is to inform drivers with real-time traffic information about the opposite connecting service by either sharing the current location of the opposite vehicle from a \textit{Vehicle Location System} (AVL), or its estimated arrival time (usually based on AVL) to the transfer site. This strategy can obviously be also applied if there is no visual contact, but it still relies on the decisions of each individual driver and their view of travel and walking time.
    \item The availability of real-time data and communication as also triggered several more recent studies \citep{Wu_ScheduleCoordination, Daganzo2017CoordinatingTime, Berrebi2015}, investigating designs and models for a centralized real-time decision support system for drivers to base their decision on.
\end{enumerate}

A recent review of seven different methods for automated decision algorithms for the holding strategy \citep{Berrebi2018} concluded that methods utilizing the prediction method achieved the best compromise between regularity and holding time.

This paper presents two novel approaches for uncertainty estimation adap\-ted and extended for the bus travel time prediction problem: \textit{Deep quantile regression} (DQR) and \textit{Bayesian recurrent neural networks} (BRNN). 
We then present a sampling technique in order to aggregate quantile estimates for link level travel time to yield the \textit{multi-link travel time distribution} needed for a vehicle to travel from its current position to a specific downstream stop point or transfer site. We show how the difference in travel time distribution for two connecting services can be used to prioritize bus holding at different strategic stop points. To demonstrate our models and approaches for uncertainty estimation, we experiment on a real-world case and dataset from a suburban area in Denmark, where two bus services are scheduled to be synchronized at a bus terminal. Based on Smart-Card ticketing data, it is indeed known that the connection is used by more than $1100$ passengers trips each month. 

At the operation level, the transfer synchronization is left up to drivers to enforce, but none of the drivers have knowledge of the arrival uncertainty of the opposite bus. How long should each driver wait at the the strategic stop points for the connecting service to allow the passenger exchange without introducing too much additional delay? If the connection assurance fails, it has severe impact on the transferring passenger's travel time. Passengers would have to wait 10, 20, or even 30 minutes for the next connection, usually forcing them to find alternatives outside of the public transport system, e.g.\ walk, request a taxi, etc. If the hold is too long, then the travel times for the non-transferring passengers are continuously increased, possibly breaking other connections downstream of the route. As suggested by \cite{Daganzo2017CoordinatingTime} one possibility would be a distributed decision support system for the drivers of each vehicle, helping them decide if they should hold at, or depart from, a strategic stop point. However, instead of assuming some constant variance of arrival time, we can use the actual predicted travel time distributions using the methods presented in this paper for quantifying the uncertainty of the specific connecting vehicle journeys. As we empirically show using real data from Copenhagen, by modelling travel time uncertainty and using a simple connection assurance technique, we can reduce bus delay by at least 33\% in the transfer scenario considered.

The rest of this paper is organized as follows: In the next section we list related work. In \Cref{sec:data_prep} we describe data requirements and preparation needed for our proposed models. In \Cref{sec:dqr} and \Cref{sec:brnn} the \textit{Deep quantile regression} (DQR) model, respectively \textit{Bayesian recurrent neural network} (BRNN) model are presented in detail including the adaption for predicting multi-link travel time distributions. In \Cref{sec:intelligent_transfer_synchronization}, we apply the \textit{uncertainty-aware} models for the task of transfer synchronization and the connection assurance application. A thorough experiment of our proposed models and adaptions is completed in \Cref{sec:experiments}, and results are presented and discussed in \Cref{sec:results}. Lastly, we conclude on our work in \Cref{sec:conclusion}.
 
\section{Literature review}

While transfer synchronization and connection assurance is a recurring topic in public transport research, modeling the travel time uncertainty for short-term and thus real time decision making is lacking from existing research. On the other  hand, short-term travel time uncertainty has been investigated separately but not related to public transport transfers, nor decision making for transfer synchronization and connection assurance.

\cite{Berrebi2015} presents a real-time bus dispatching policy, and models uncertainty but only based on a fixed variance of travel time for each link (i.e.\ homoscedastic uncertainty). In contrast, our methods use recent link travel time observations to quantify and estimate the short-term travel time uncertainty, i.e.\ heteroscedastic uncertainty.

\cite{Rodr2018} presented a deep  quantile regression (DQR) model and evaluated on traffic speed uncertainty among other datasets. Similar to this work, it used a convolutional and recurrent neural network architecture, and showed this architecture was able to obtain the tightest prediction intervals and still be covering as expected. It also showed that DQR generally outperformed other popular methods for uncertainty estimation on the evaluated datasets, including Monte Carlo dropout \citep{MCDropOut} and linear quantile regression \citep{Koenker2001}. Our research extends this path, by comparing DQR with Bayesian recurrent neural networks (BRNN), and use the predicted travel time uncertainty estimates for real-time decision making in transfer synchronization.

\cite{Petersen2019} presented a convolutional and recurrent neural network architecture for  point-prediction of bus travel time, including multiple time-steps into the future. They  showed its ability to capture some of the spatio-temporal correlations that are naturally present in a public transport network. We have used this as inspiration and extended this in our work with an uncertainty-aware network architecture using respectively DQR and BRNN.

Finally \cite{OSullivan2016} presented a meta-model approach that augments fixed sampled bus arrival point-predictions from existing AVL systems for a single stop point. By using Gaussian processing quantile regression a bound is placed on the expected error of the point-prediction, and they present arrival error estimates with heteroscedastic uncertainty. Our approach differs from this, since we predict multiple time-steps ahead in time, each with its own uncertainty estimate as a direct output of our models. This removes the overhead of fitting an additional meta-model completely, which constitute a significantly computationally effort, especially given the use of Gaussian processing. Furthermore, since we model uncertainty for all stops simultaneously, we are able to capture to spatio-temporal correlations. This together allows a more elegant and accurate use for connection assurance, where the time until the connection might by multiple time-steps into the future, and the path between the current location of the vehicles and the transfer site usually consists of multiple stops. 

\section{Methodology}
In this section, we present our two proposed models for link travel time prediction with uncertainty estimation: \textit{Deep quantile regression} (DQR) and \textit{Bayesian recurrent neural networks} (BRNN), and we propose an approach for obtaining uncertainty estimates for vehicle run time across multiple links, e.g.\ from current location to the transfer site. We finalize by demonstrating the application of this methodology to the connection assurance and bus holding strategy.

\subsection{Data requirements and preparation}
\label{sec:data_prep}
For both the DQR and the BRNN model, we assume input for each time step $t \in \{1, \ldots, N\}$ of the form $\mathbf{X}_t \in \mathbb{R}^{U \times N_{\textit{ln}}}$, where $U$ is the window size of historic travel time observations, and $N_{\textit{ln}}$ is the number of links included in the model. To align the multiple irregular time series of link travel time observations to the regular reference time ($t \in \{1, \ldots, N\}$), we impose a similar technique to \cite{Petersen2019}, which snaps the observations to a fixed frequency reference time, e.g.\ 15min time steps. For training the network, we further assume $\mathbf{Y}_t \in \mathbb{R}^{K \times N_{\textit{ln}}}$, where $K$ is the number of time steps to predict into the future at time step $t$. Hence, $\mathbf{X}_t$ and $\mathbf{Y}_t$ can be seen as multivariate fixed frequency time series folded onto themselves along the temporal axis $U$ and $K$ times, respectively. However, the transformation introduces the need for imputation, since not all links are guaranteed to have been observed in all $[t,t+1]$ intervals. We suggest imputing using a per-link mean conditioned on some recurring features, e.g. time of day and day of week: $\text{E}\left[\textbf{X}_t^\textit{ln} \, \middle| \, \text{DoW}(t), \text{ToD}(t) \right]$. The choice of the frequency of the reference time should be done to minimize the need for imputation, while data masking can minimize the contribution of the imputed values to training of the network.

\subsection{Deep quantile regression}
\label{sec:dqr}

Neural network quantile regression is a simple, yet elegant extension to classic neural network regression. Neural network regression estimates the expected value of one or more continuous variables. For point estimates the most popular loss function is the sum of squares loss, or $l_2$ loss cf.~\cref{eq:l2_loss}, where $\mathcal{D}$ is the set of indices for all elements in the true and estimated tensors.
\begin{equation}
  l_2(\mathbf{Y}, \hat{\mathbf{Y}}) = \sum_{d \in \mathcal{D}} \left( \mathbf{Y}_d - \mathbf{\hat{Y}}_d \right)^2  
  \label{eq:l2_loss}
\end{equation}
Similarly, we can optimize the network parameters for a quantile loss, $l_q^p$, cf.~\cref{eq:quantile_loss} given a specific quantile $p$.

\begin{equation}
l^p_q(\mathbf{Y},\hat{\mathbf{Q}}^{p}) = \sum_{d \in \mathcal{D}} \max \left( p \cdot \left(\mathbf{Y}_d - \hat{\mathbf{Q}}^{p}_d \right), (1-p) \cdot \left(\mathbf{Y}_d - \hat{\mathbf{Q}}^{p}_{d} \right) \right)
\label{eq:quantile_loss}
\end{equation}

\cite{Rodr2018} showed that both $l_2$ and multiple quantile losses can be estimated jointly using the loss cf.\ \cref{eq:joint_loss}.
\begin{align}
 \small
\begin{split}
     \mathcal{L}(\mathbf{Y},\hat{\mathbf{Y}} ,\hat{\mathbf{Q}}^{p_1}, \dots, \hat{\mathbf{Q}}^{p_J} ) &= \sum_{d \in \mathcal{D}}  \biggl( \left( \mathbf{Y}_d - \mathbf{\hat{Y}}_d \right)^2  \\ 
    &+ \sum_{j=1}^{J} \max \left( p_j\cdot \left(\mathbf{Y}_d - \hat{\mathbf{Q}}^{(p_j)}_d \right), (1-p_j)\cdot \left(\mathbf{Y}_d - \hat{\mathbf{Q}}^{(p_j)}_d \right) \right)  \biggr)
    \label{eq:joint_loss}
\end{split}    
\end{align}
When quantile regression is applied in the context of deep neural networks, i.e.\ multiple blocks of non-linear layers, we refer to the technique as deep quantile regression (DQR).

\subsubsection{DQR link travel time prediction model}
The link travel time model used for deep quantile regression is similar in network architecture to the mean prediction model by \cite{Petersen2019}, who utilized
\textit{Long short-term memory} (LSTM) cells in combination with 1D convolutions over the links. This approach showed to be capable of capturing spatio-temporal correlations. 

\begin{figure}[t!]
    \centering
    \includegraphics[width=\textwidth]{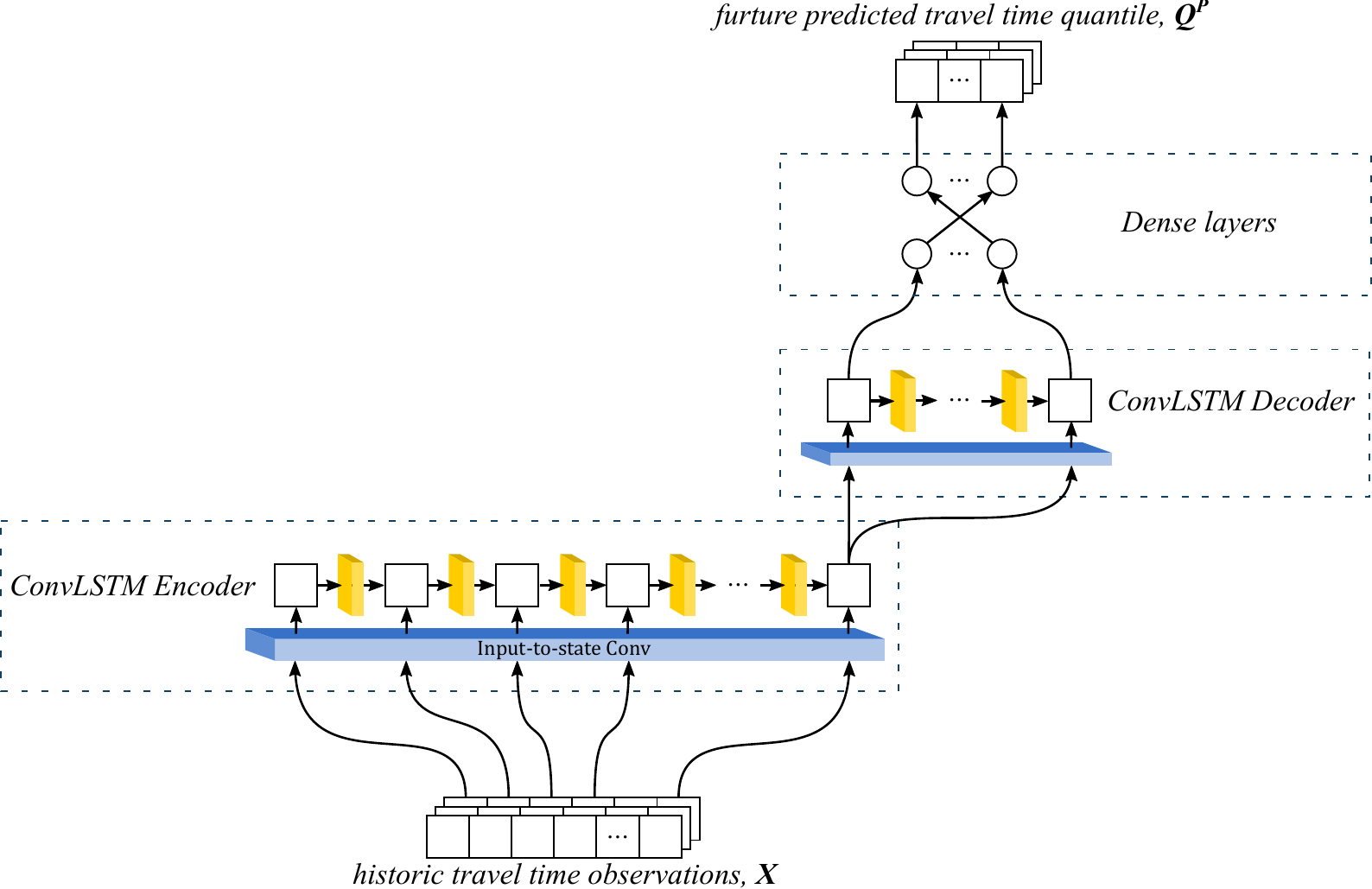}
    \caption{Network architecture for the DQR ConvLSTM link travel time model.}
    \label{fig:DQRModel}
\end{figure}

The adapted network architecture is illustrated in \Cref{fig:DQRModel}, where the blue blocks illustrates the 1D input convolutions, and yellow blocks the 1D state-to-state convolutions between LSTM cells. We have also adapted the loss to the quantile regression task cf.~\cref{eq:joint_loss}, and reduced the stacking of recurrent blocks, such that we have a single-layer encoder and a single-layer decoder. The hyper-parameters for the network include the \textit{Convolutional Kernel Size}, the \textit{LSTM Hidden State} and the \textit{Dropout Probability}. In \Cref{sec:hp}, we detail an approach for choosing the hyper-parameters. We further standardize the input and output of the DQR neural network by mean centering and scaling $\hat{\textbf{X}}_t^\textit{ln}$ and $\hat{\textbf{Y}}_t^\textit{ln}$. We use the same per-link conditional mean as used for imputation and its corresponding variance cf.~\Cref{eq:standarize}. This ensures that input values are centered around zero, which is appropriate for the activation function of the LSTM-cells (tanh).

\begin{equation}
\bar{\textbf{X}}_t^\textit{ln} = \frac{\textbf{X}_t^\textit{ln} - \text{E}\left[\textbf{X}_t^\textit{ln} \, \middle| \, \text{DoW}(t), \text{ToD}(t) \right]}{\sqrt{\text{Var}\left[\textbf{X}_t^\textit{ln} \, \middle| \, \text{DoW}(t), \text{ToD}(t) \right]}}
    \label{eq:standarize}
\end{equation}

For the task of bus holding and connection assurance, we are not only interested in the per-link predicted quantiles, but we also need the accumulated uncertainty estimate of the travel time between the current location of the vehicle and the transfer site. While the predicted means can be easily summed across links, the quantiles cannot. Our approach around this is to sample from the travel time distribution for each link, and accumulating (i.e.\ sum) the samples. This follows the assumption that the uncertainty is independent across each link, which generally holds for shorter road segments such as stop-to-stop bus links. The aggregated sampled link travel time yields a probability distribution for the total travel time, and thus a probability distribution for the arrival at the transfer site.

The predicted mean, $\hat{\textbf{Y}}_t^\textit{ln}$, and the predicted quantiles, $\hat{\mathbf{Q}}_t^{\textit{ln}, p_1}, \dots, \hat{\mathbf{Q}}_t^{\textit{ln}, p_J}$, for link $ln$ at time step $t$, are directly available as output from the DQR model. Since the mean is already estimated by the joint model directly, we only need to estimate the scale, $\sigma^{\textit{ln}}_t$, in order to sample from the link travel time distribution cf.~\cref{eq:curve_fit}.
\begin{equation}
    \hat{\textbf{S}}^{\textit{ln}}_t \sim \mathcal{N}(\hat{\textbf{Y}}^{\textit{ln}}_t,  \left[ \sigma^{\textit{ln}}_t \, \middle| \, \hat{\mathbf{Q}}^{\textit{ln}, p_1}_t, \dots, \hat{\mathbf{Q}}^{\textit{ln}, p_J}_t, p_1, \dots, p_J \right])
    \label{eq:curve_fit}
\end{equation}

\begin{figure}[t]
    \centering
    \includegraphics[width=\textwidth]{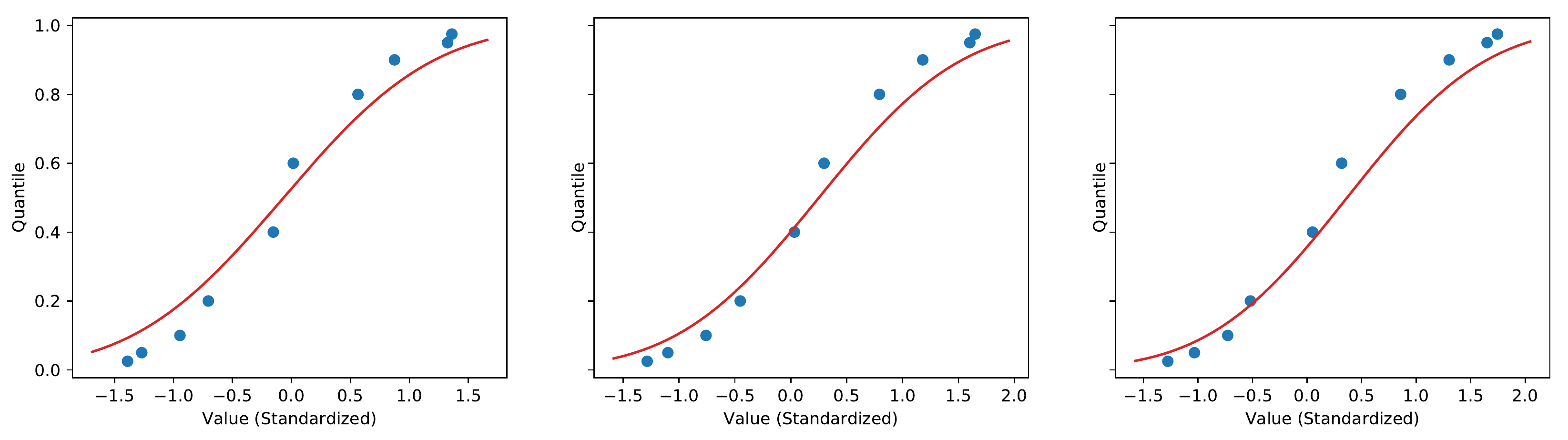}
    \caption{Three examples of sampling from DQR using curve fitting. The blue dots are the predicted quantiles, and the red line the fitted CDF for $X^{\textit{ln}}_t \sim \mathcal{N}(\hat{\textbf{Y}}^{\textit{ln}}_t,  \sigma^{\textit{ln}}_t)$ }
    \label{fig:curve_fit}
\end{figure}

We propose to estimate $\sigma^{\textit{ln}}_t$ using least squares optimization (e.g.\ LM algorithm) by minimizing the error in $p_i = \text{P} (X < \hat{\mathbf{Q}}_t^{\textit{ln}, p_i} \, | \, X \sim \mathcal{N}(\hat{\textbf{Y}}^{\textit{ln}}_t,   \sigma^{\textit{ln}}_t) + \epsilon )$ for all $i \in \{1, \dots, J\}$. \Cref{fig:curve_fit} shows three examples of fitted CDFs from DQR using LM fitting. Even though the LM fitting is computationally expensive compared to the predictive call of the DQR model, we argue that it is indeed feasible to do in real-time since the fitting is only needed once for each link prediction. At this point, we have achieved the goal of predicting the uncertainty of arrival at the transfer site, since we can computationally inexpensively draw a significant number of samples and summarize them by $\hat{\textbf{R}}_{\textit{L},t} = \sum_{\textit{ln} \in \textit{L}} \hat{\textbf{S}}^{\textit{ln}}_t$ for all relevant links, $\textit{ln}$, thus giving an uncertainty estimate for the travel time across all the links.

\subsection{Bayesian recurrent neural networks}
\label{sec:brnn}

Bayesian neural networks (BNN) \citep{Blundell2015} provide a fundamentally different approach for handling uncertainties. In ``standard'' neural networks, each weight in the network is estimated as a point estimate during training. On the contrary, in Bayesian neural networks, the weights are estimated as probabilistic distributions. Weights with greater uncertainty introduce more variability into the downstream network layers, and thus eventually to the output. Bayesian recurrent neural networks (BRNN) \citep{BRNN} is the adoption of this probabilistic approach for weights estimation onto the classic recurrent neural networks (RNN). We refer to \cite{Blundell2015} and  \cite{BRNN} for a complete in-depth review, including propositions and prior assumptions. However, we present essential aspects, for those not familiar with the prior work, including further adjustments and refinements that we have contributed. 

\subsubsection{Variational inference}
Variational inference is an approximate inference method where variational parameters $\boldsymbol{\nu}$ are optimized in order to make a variational distribution  $q(\boldsymbol{w} \mid \boldsymbol{\nu} )$ as close as possible to the true posterior $P(\boldsymbol{w} \mid \mathbf{X}, \mathbf{Y})$. A family of tractable distributions is chosen for $q$ and thus the approximate posterior $q(\boldsymbol{w} \mid \boldsymbol{\nu} )$ can be used to make predictions for unseen data. The Kullback-Leibler divergence ($\mathbb{KL}$) is used to quantify closeness between two distributions, and thus the goal is to find the variational parameters $\boldsymbol{\nu}$ that minimise the KL-divergence cf.\ \cref{eq:kl_arg_min}. 
\begin{equation}
    \boldsymbol{\nu}^* = \arg \min_{\boldsymbol{\nu}} \mathbb{KL}\left( q(\boldsymbol{w} \mid \boldsymbol{\nu}) \mid P(\boldsymbol{w} \mid \mathbf{X}, \mathbf{Y}) \right)
    \label{eq:kl_arg_min}
\end{equation}
A cost function achieving this is the \textit{evidence lower bound} (ELBO) cf. \cref{eq:elbo}. 

\begin{align}
\mathcal{L}_\text{ELBO}(\boldsymbol{\nu}) &=  \mathbb{KL}(q(\boldsymbol{w} \mid \boldsymbol{\nu}) \mid P(\boldsymbol{w}))- \sum_{t=1}^{N} \int_{\boldsymbol{w}} q(\boldsymbol{w} \mid \boldsymbol{\nu}) \log P(\mathbf{Y}_t \mid \boldsymbol{w}, \mathbf{X}_t)
    \label{eq:elbo}
\end{align}
The first part is the KL-divergence between the variational distribution and the prior over the weights, which acts as a regularization term. The second part is the expectation of the log-probability of the observed data, $\log P(\mathbf{Y} \mid \boldsymbol{w}, \mathbf{X})$, under the variational distribution. Since we are dealing with log-probabilities, the second term becomes a sum over all the data points. For scalability purposes, it is desirable to train the network using mini-batches. We can divide the data into $B$ equal sized batches, each with batch size $N_B$, and each mini-batch is then contributing with loss cf. \cref{eq:elbo_b}. 
\begin{align}
\mathcal{L}^{(b)}_\text{ELBO}(\boldsymbol{\nu}) &=  \frac{1}{B}\mathbb{KL}(q(\boldsymbol{w} \mid \boldsymbol{\nu}) \mid P(\boldsymbol{w}))- \sum_{i=1}^{N_B} \int_{\boldsymbol{w}} q(\boldsymbol{w} \mid \boldsymbol{\nu}) \log P(\mathbf{Y}^{(b)} \mid\boldsymbol{w}, \mathbf{X}^{(b)}) \label{eq:elbo_b} 
\end{align}

When training, we rely on Monte Carlo sampling with $N_\textit{MC}$ samples of $\boldsymbol{w}^{(s)}$ from the posterior for approximating the gradients of the integral $\int_{\boldsymbol{w}} q(\boldsymbol{w} \mid \boldsymbol{\nu}) \log P(\mathbf{Y}^{(b)} \mid\boldsymbol{w}, \mathbf{X}^{(b)})$ and the KL-term. For recurrent neural networks, as is the case with our model, $\boldsymbol{w}$ is shared between all cells in a recurrent layer (e.g.\ LSTM layer), and the input is the combination of the hidden state of the previous cell, $\mathbf{h}_{u-1}$, and the $u$'th input of the sequence input at time step $t$, $\mathbf{x}_{t,i}$ as shown in \cref{eq:elbo_rnn}. Here, $\boldsymbol{f}_h$ and $\boldsymbol{f}_y$ and are the \textit{state-to-state} transition function and the \textit{state-to-output} function of the recurrent layer, respectively.

\begin{align}
    \mathcal{L}_\text{ELBO,RNN}^{(b)}(\boldsymbol{\nu}) \approx &\frac{1}{B}\left(\log q(\boldsymbol{w}^{(s)} | \boldsymbol{\nu}) - \log P(\boldsymbol{w}^{(s)})\right) \nonumber \\ &-\sum_{i=1}^{N_B} \log P(\mathbf{y}^{(b)}_{u} \mid \boldsymbol{f}_y (\boldsymbol{f}_h (\mathbf{x}^{(b)}_{u}, \mathbf{h}_{u-1})))
\label{eq:elbo_rnn}
\end{align}

\subsubsection{Training, prior and posterior choices}

The model can be trained using the \textit{Bayes by Backprop} algorithm \citep{Blundell2015, BRNN}: For every iteration of optimisation, every network parameter $w_m$ with variational parameters $\boldsymbol{\nu}_m = (\mu_m, \rho_m)$ is updated by sampling $\epsilon \sim \mathcal{N}(0,1)$ and setting $w_m = \mu_m + \sigma_m \epsilon$, where $\sigma_m = \log(1 + \exp(\rho_m))$, then the loss cf. \cref{eq:elbo_b,eq:elbo_rnn} is calculated and we perform back-propagation as usual.

    For the prior over the weights, we suggest the \textit{scale mixture prior} as proposed by \cite{Blundell2015} and shown in \cref{eq:mixture_prior}, where  $\mathcal{N}(x \mid \mu, \sigma^2)$ denote the univariate normal probability density function. When $\sigma_2 \ll \sigma_1$ the \textit{scale mixture prior} can be seen as probabilistic \textit{dropout}, i.e.\ weights have a probability $1 - \pi$ of having to be close to zero to satisfy the KL-regularization term in \cref{eq:elbo,eq:elbo_rnn}. Thus $\pi, \; \sigma_1, \; \sigma_2$ are hyper-parameters for the network to be tuned for the specific task and  variability in the data, which we will return to in \cref{sec:hp}.
\begin{equation}
    P(\boldsymbol{w}) = \prod_m \left(\pi \mathcal{N} (w_m \mid 0, \sigma^2_{1}) + (1-\pi) \mathcal{N} (w_m \mid 0, \sigma_2^2)\right)
\label{eq:mixture_prior}
\end{equation}
Lastly, the variational distribution, from which we will sample from to propagate uncertainty to downstream layers is given by \cref{eq:variational_postrior}:
\begin{equation}
q(\boldsymbol{w}_m \mid \boldsymbol{\nu}) = \mathcal{N}(\boldsymbol{w}_m \mid \mu_m, \sigma^2_m)
\label{eq:variational_postrior}
\end{equation}

\subsubsection{BRNN link travel time prediction model}

\begin{figure}[t!]
    \centering
    \includegraphics[width=\textwidth]{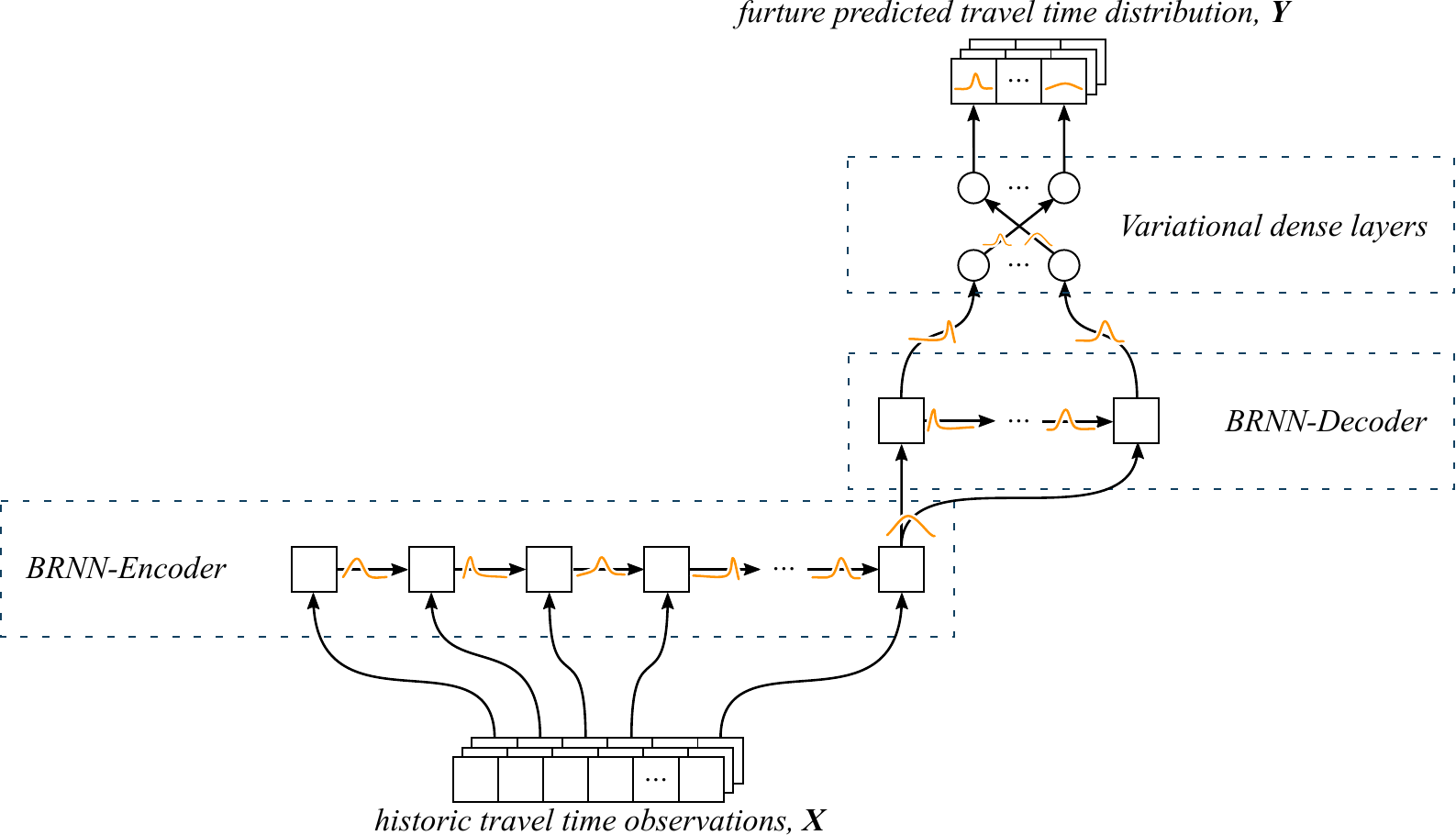}
    \caption{Network architecture for the BRNN link travel time model.}
    \label{fig:BRNNModel}
\end{figure}

Figure~\ref{fig:BRNNModel} shows the overall architecture of the proposed \textit{BRNN link prediction model}. Each square in the BRNN-encoder and BRNN-decoder blocks represents a \textit{Bayesian LSTM Cell}. Overall, the structure is similar to the DQR model, but without the convolutional transformations and thus without \textit{weight sharing} along the spatial dimension. 
We also tested a BRNN model with convolutional layers, but were unable to achieve satisfactory results. 

Since the uncertainty is embedded directly into the weights of the \textit{BRNN link prediction model}, we can simply sample from the network without additional computational costs. As such, the BRNN model can been seen as an ensemble of link travel time models: each time we input the same historic window of link travel times, the result will be varying as a result of the uncertainty learned into the weights of the network given the state represented by the historic window.  
At this point, we can then computationally inexpensively draw a significant number of samples, $\hat{\textbf{Y}}_t^{\textit{ln}}$, from the BRNN model, and summarize them by $\sum_\textit{ln} \hat{\textbf{Y}}_t^{\textit{ln}}$ for all relevant links, $\textit{ln}$, thus giving an uncertainty estimate for the travel time across all the links.

\subsection{Transfer synchronization with intelligent uncertainty handling}
\label{sec:intelligent_transfer_synchronization}

In the previous sections, we have presented two distinct approaches for sampling short-term link travel time for multiple time steps into the future. Although different in terms of modelling uncertainty, they consist of a similar and established network architecture for link travel time prediction, and are both capable of quantifying the travel time uncertainty for each link. We have presented a simple aggregation of the output of the models, such that travel time uncertainty can be predicted from the current position of a vehicle to an arbitrarily stop point downstream on its route. We will now apply the \textit{uncertainty-aware} models for the task of transfer synchronization and the connection assurance application as described in \Cref{sec:intro}.

As $\hat{\textbf{R}}_{\textit{L},t} = \sum_{\textit{ln} \in \textit{L}} \hat{\textbf{S}}^{\textit{ln}}_t$ constitute point estimates based a sum of samples, $\textbf{R}_{\textit{L},t}$ can be considered as a distribution itself, namely the travel time distribution for a relevant set of links, $L$, between the current position of the vehicle and the transfer site, whose expected-value is approximated by $\hat{\textbf{R}}_{\textit{L},t}$. We do this for both of the vehicles, \textit{Bus 1} and \textit{Bus 2}, included in the connection assurance. When sampling from $\textbf{R}_{\textit{L},t}$, special care needs to be taken for cases where the accumulated link travel time exceeds the regular time frequency of the models. In this case, samples should be drawn to capture this for the rest of the summation by sampling from the relevant time horizon output (e.g. $t + 1$, $t + 2$, etc.) and continuing. We denote a sample drawn for \textit{Bus 1} as $r_1 \sim {\textbf{R}}_{\textit{L}_1,t}$ and  $r_2 \sim {\textbf{R}}_{\textit{L}_2,t}$ for\textit{Bus 2}, where $L_1$ and $L_2$ are the sets of links to be traversed by \textit{Bus 1}, respectively \textit{Bus 2}.

We define the distribution of the difference in travel time as \cref{eq:connection_prop}, i.e. a distribution of time that can be prioritized for bus holding at strategic points. In section \Cref{sec:decision_support_system}, we demonstrate how this probability distribution can be used in a decision support system for minimizing sub-optimal delays in accordance with the connection assurance application.
\begin{equation}
    P\left( (r_1 - r_2) \mid r_1 \sim {\textbf{R}}_{\textit{L}_1,t}, \; r_2 \sim {\textbf{R}}_{\textit{L}_2,t} \right) 
    \label{eq:connection_prop}
\end{equation}

\section{Experiments}
\label{sec:experiments}

In this section, we present a case study to evaluate the two presented models for link travel time uncertainty estimation, and the sampling technique to yield vehicle travel time distribution across multiple links.   

We used Smart-Card ticketing data to find interesting transfer sites in the Greater Copenhagen Area, Denmark. We weighted the number of transfers with a penalty linear to the headway of the connecting service, such that we consider both connections used by many passengers, but also transfers where passengers face a long wait time if the connection is missed. We selected the following case study as the one best matching the criteria above.

\begin{figure}[t]
    \centering
    \includegraphics[width=8cm]{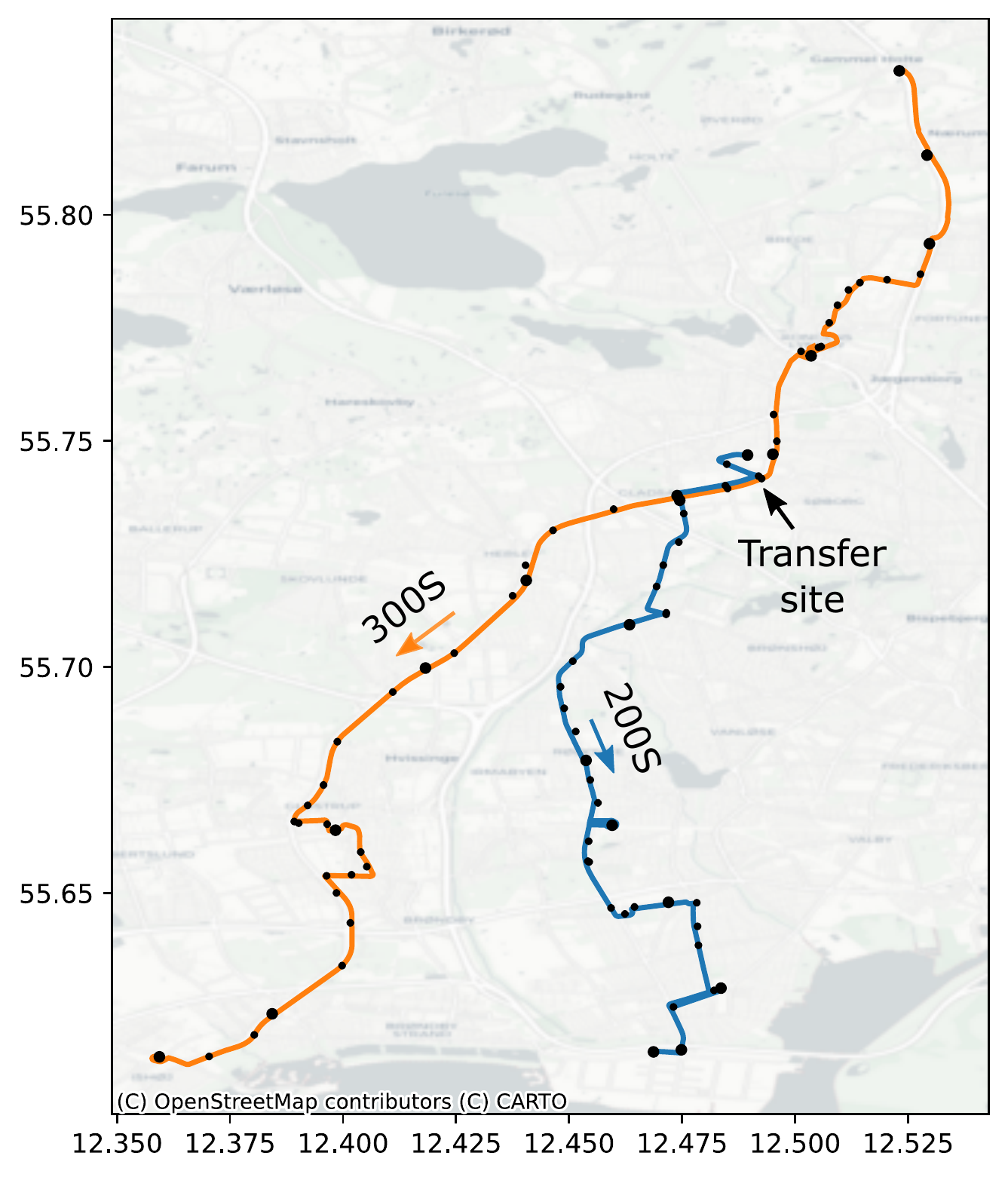}
    \caption{Map of case study lines and transfer site. 300S (orange) and 200S (blue). Basemap credits: OpenStreepMap, CARTO.}
    \label{fig:map}
\end{figure}

\subsection{Case study area}
\Cref{fig:map} shows the area of the selected case study, which consist of the transfer site between two major express bus lines in the Copenhagen suburbs. Bus line 300S (south-west bound) is connecting with bus line 200S (southbound) at a transfer site approximately 1/3 into the route of 300S. The 200S line is not far into its route at this site, but the site is not visible from the origin of the line. A lot of passengers that have boarded the 300S line in the north transfer to the 200S line at this site, going more directly south.  In October 2020, more than 1.100 transfers from 300S to 200S were recorded at this site by the Smart-Card ticketing system. It is worth noticing, that the Smart-Card ticketing system is only used by approximately 30\% of the passengers. The remaining share of passenger uses mainly monthly passes or cash tickets, for which transfers are not easily tractable. Thus it is expected that the true number of monthly passenger transfers at the transfer site used in this case study is above 3.000.

\begin{figure}[t]
    \centering
    \includegraphics[width=8cm]{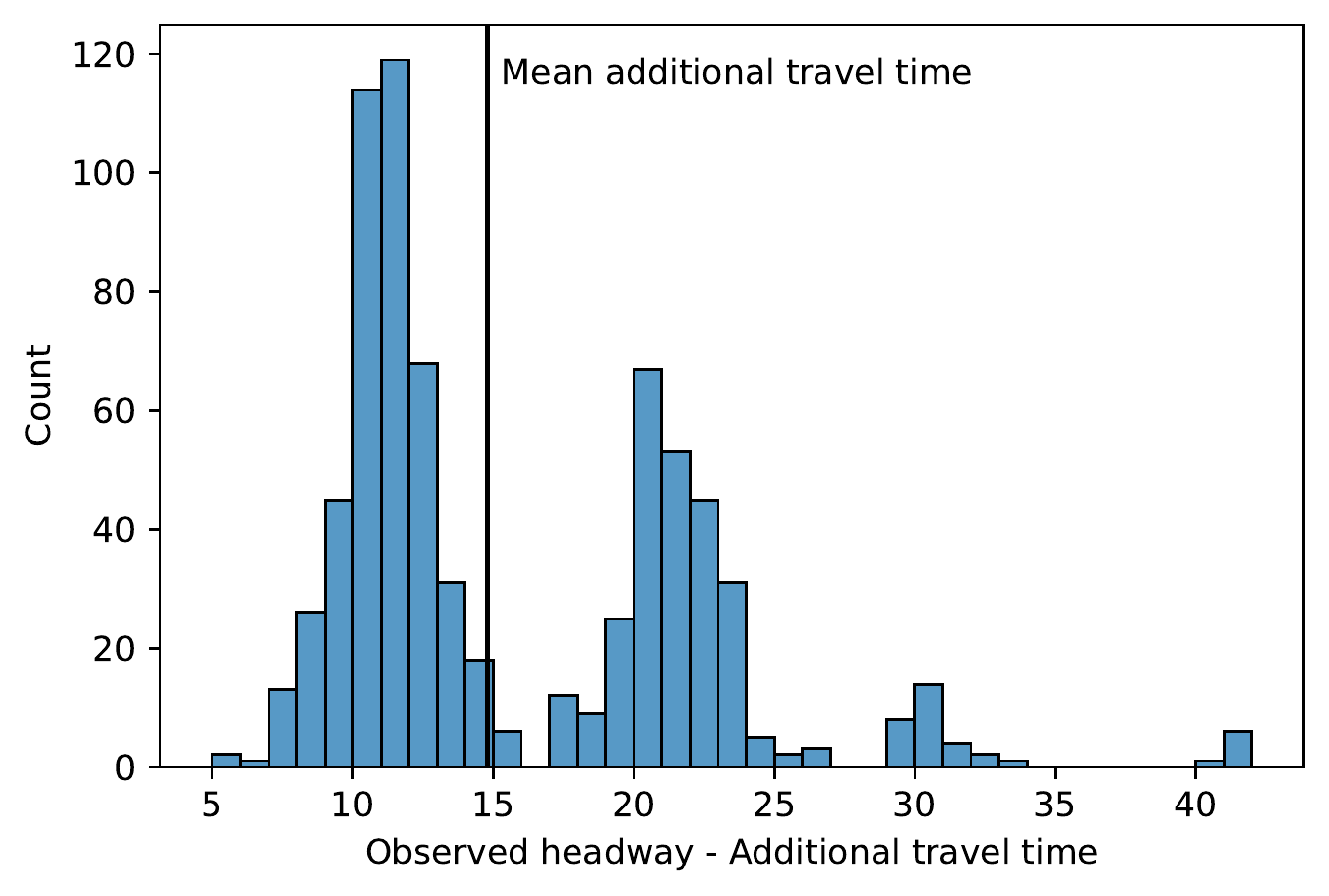}
    \caption{Observed headway of 200S for observed transfers.}
    \label{fig:headway}
\end{figure}

We calculated the expected headway for cases where passengers missed their connections, i.e.\ how much additional travel time one should expect. The distribution is shown in \Cref{fig:headway}. The peaks corresponds to different scheduled frequencies of the 200S line, and it is clear that even though a lot of passengers \textit{only} will be delayed between 5 and 15 minutes, there is a long tail, with a mean of additional travel time of approximately $15$ minutes. As such, 66\% of the passengers would have been delayed more than the mean of approximately $15$ minutes. This supports the need for a bus holding strategy and a decision support system that intelligently handles the uncertainty of travel time present in the bus system.

\subsection{Link travel time dataset}
We use link travel time data collected for 200S and 300S for the period of August 3, 2020 to November 29, 2020, corresponding to 17 full weeks. In total, more than $500\,000$ link travel time observations are used in the case study. We divide the period into a training set (first 13 weeks), a validation set used for hyper-parameter tuning (next 2 weeks), and a test data set used for evaluation and results (last 2 weeks). \Cref{tab:dataset} shows a summary of the data available for each of the lines in the case study. For each bus line, we arrange the data in accordance with \Cref{sec:data_prep}, which yields three distinct sets of $\mathbf{X}_t$ and $\mathbf{Y}_t$ for training, validation, and testing, respectively. We choose a regular frequency of 15min for the reference time based on the headway distribution shown in \Cref{fig:headway}. Similar to \cite{Petersen2019}, we chose a historic window size of $w = 32$ (8 hours) and a prediction horizon of $k = 3$ (45 minutes). For baseline comparison, we have further included a \textit{Linear Kalman filter} using a state-space corresponding to the number of links. 

\begin{table}[t]
    \centering
    \begin{tabular}{c|cc }
          & 200S & 300S   \\\hline \hline 
         Link count  & 29  & 39 \\ \hline 
         Link travel time observations & $218\,956$  & $333\,118$ \\ \hline 
         Train: $\mathbf{X}_t$ size & $8701 \times 32 \times 29$ & $8701 \times 32 \times 39$ \\  \hline 
         Train: $\mathbf{Y}_t$ size & $8701 \times 3 \times 29$ & $8701 \times 3 \times 39$ \\  \hline 
         Validation: $\mathbf{X}_t$ size & $1309 \times 32 \times 29$ & $1309 \times 32 \times 39$ \\ \hline
         Validation: $\mathbf{Y}_t$ size & $1309 \times 3 \times 29$ & $1309 \times 3 \times 39$ \\ \hline
         Test: $\mathbf{X}_t$ size & $1309 \times 32 \times 29$ & $1309 \times 32 \times 39$ \\ \hline 
         Test: $\mathbf{Y}_t$ size & $1309 \times 3 \times 29$ & $1309 \times 3 \times 39$ \\ \hline
    \end{tabular}
    \caption{Summery of case study dataset}
    \label{tab:dataset}
\end{table}

\subsection{Implementation and reproducibility}
We have implemented the DQR model using \textit{Keras} \citep{Keras} and \textit{TensorFlow} \citep{TensorFlow}, while the BRNN model is implemented with \textit{PyTorch} \citep{NEURIPS2019_9015}, but also builds on a Bayesian extension for PyTorch by \cite{esposito2020blitzbdl} called \textit{Blitz}. We have published the source code and data for the reproducibility of our results in its entirety at: \url{https://github.com/MLSM-at-DTU} 

\subsection{Hyper-parameter tuning}
\label{sec:hp}

Both the DQR and the BRNN model (and baselines) rely on carefully choosing a number of hyper-parameters. We tuned the latter based on the train and validation data sets. Specifically for our proposed models, we utilized Bayesian Optimization with Gaussian Processes \citep{GpMinimize_Snoek2012}, since our search space, as summarized in \Cref{tab:hp_search_space}, is impractical to investigate with traditional grid search techniques. The Kalman filter baseline parameters (e.g.\ transition- and observation covariance matrices, and initial state) was estimated using the \textit{Expectation-Maximization Algorithm} \citep{EmAlgorithm}.
\begin{table}[t]
    \centering
    \begin{tabular}{c|cc|cc}
          & \multicolumn{2}{c|}{DQR model} & \multicolumn{2}{c}{BRNN model} \\
          & & & & \\[-8pt]
          & Lower  &  Upper& Lower  & Upper  \\
         Parameter & bound &  bound &   bound &   bound \\        
         \hline \hline
         LSTM State Size & 10 & 128 & 10 & 50  \\ \hline
         ConvLSTM Kernel Size & 1 & 20 & - & -  \\ \hline
         Dropout Probability & 0.0 & 0.6  & - & - \\ \hline
         Mixture Probability, $\pi$ & - & - & 0.7 & 1.0 \\ \hline
         Mixture Prior, $\sigma_1$ & - & - & 1 & 3 \\ \hline
         Mixture Prior, $\sigma_2$ & - & - & 0.001 & 1 \\ \hline
    \end{tabular}
    \caption{DQR and BRNN hyper-parameter search space}
    \label{tab:hp_search_space}
\end{table}

\section{Results and discussion}
\label{sec:results}

On the test dataset, we evaluated the presented models' ability to: (i) predict link level travel time uncertainty, (ii) multi-link travel time uncertainty using the proposed aggregation technique, and finally (iii) the performance when applied to the task of \textit{transfer synchronization} and \textit{connection assurance}. We draw 500 samples from each model for each link, for each time step predicted in the test data set. Thus the sampled model output from each of the two models has size
$500 \times 1309 \times 3 \times 29$ for 200S and $500 \times 1309 \times 3 \times 39$ for 300S.

\subsection{Single link travel time the performance}

\begin{figure}[t]
    \centering
    \includegraphics[width=\textwidth]{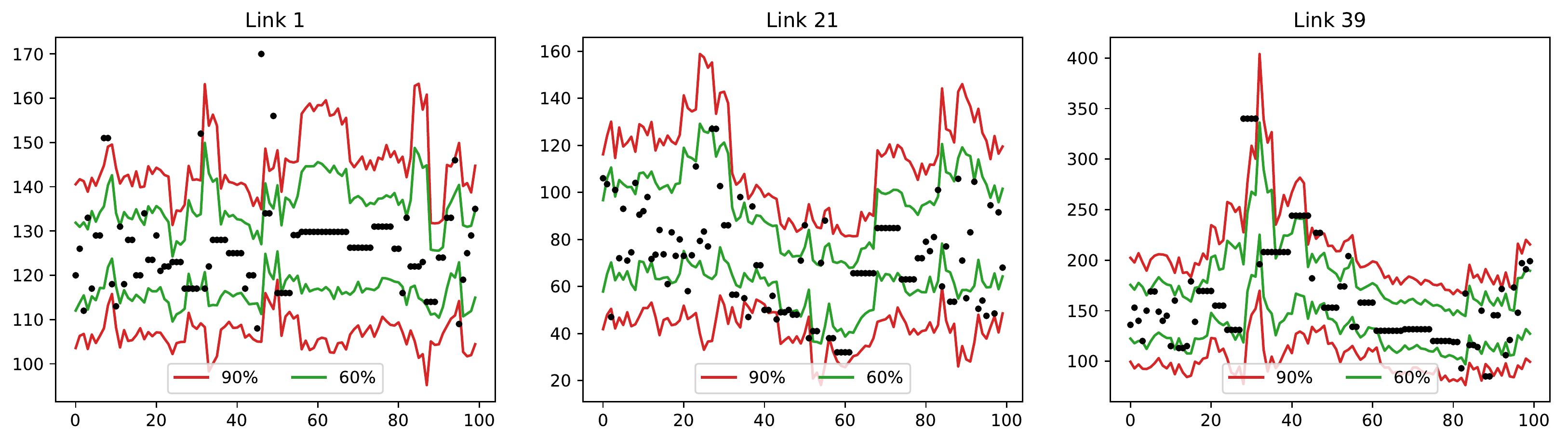}
    \caption{DQR results from three random links from the 300S line}
    \label{fig:dqr-quantiles}
\end{figure}

\begin{figure}[t]
    \centering
    \includegraphics[width=\textwidth]{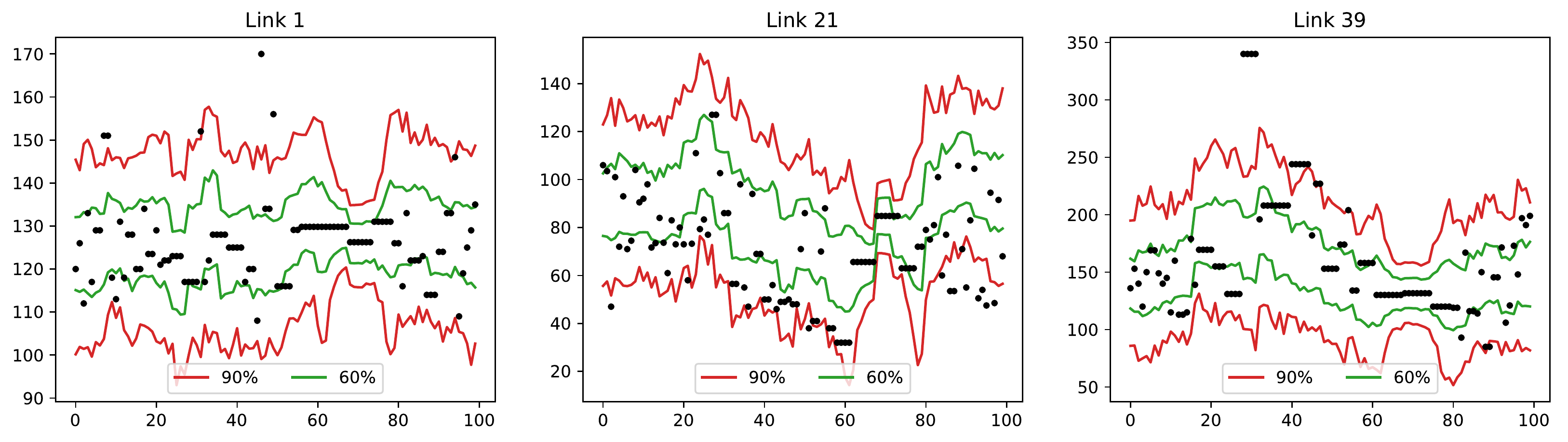}
    \caption{BRNN results from three random links from the 300S line}
    \label{fig:brnn-quantiles}
\end{figure}

For the link level travel time uncertainty, \Cref{fig:dqr-quantiles,fig:brnn-quantiles} show the model output for two selected prediction intervals (60\% and 90\%) for three randomly chosen links at $t + 1$ on the 300S line for subset of the test period. We note some interesting differences: for Link 1 and Link 21 the DQR model is not able to contract the uncertainty interval between time steps 60 and 80, which are during the night hours, even though very little link travel time variability is observed. On the contrary, the BRNN model is able contract the uncertainty estimation at this point in time. On the other hand, the DQR model captures the uncertainty with respect to the peaks for Link 39 around time step 30, which is mostly missed by the BRNN model. This indicates that both models have strengths and weaknesses, which we return to discuss in \Cref{sec:conclusion}.

\subsection{Multi-link travel time the performance}

For the multi-link travel time, the performance of the models is investigated the following metrics:
\begin{itemize}
    \item \textit{Interval coverage percentage} (ICP),  which, for a given uncertainty interval, measures a model's ability to empirically cover that interval. An ICP value close to the expected uncertainty interval indicates a model that neither over- nor underestimates the uncertainty.
    \item \textit{Mean interval length} (MIL), which, for a given uncertainty interval, measures the average length of the prediction intervals. Smaller MIL values are desired given a reasonable ICP value indicating that the predicted uncertainty is as low as possible.
    \item The point prediction accuracy using the \textit{root mean square error} (RMSE) cf. $\sqrt{l_2(\mathbf{Y}, \hat{\mathbf{Y}})}$) and \cref{eq:l2_loss}. For the DQR model we directly have a point estimate, while for the BRNN model we use the mean of the sampled link travel times as a point estimate.
\end{itemize}

Furthermore, we also evaluate the different models across different prediction horizons: 15, 30 and 45 minutes into the future. \Cref{tab:DQR_results} and \Cref{tab:brnn_results} show the result metrics for the multi-link travel time uncertainty using the proposed aggregation technique for all 39 links of the 300S bus line. For comparison, we have included the same metrics for the Kalman filter baseline in \Cref{tab:kalman_results}.  

Overall, we observe the best performance for point prediction with the DQR model, both compared to the BRNN and the Kalman filter, which in fact also is outperforming the BRNN model with respect to point prediction for $t + 1$ and $t + 2$ (15 and 30 minute into the future). We attribute this result to the dedicated output for the point prediction of the DQR models, in contrast to both the BRNN and the Kalman filter, where the point prediction is the mean of the samples. For skewed uncertainty distributions, DQR has the ability to adjust the point prediction independent of the predicted quantiles. We observe a consistent drop in point prediction accuracy when predicting further into the future (e.g. 30 and 45 minutes) for the DQR model and the Kalman filter, which is expected. In contrast, the BRNN maintain around the same, but higher, prediction error, which indicates BRNN is not able to learn the time-dependent error with respect to the prediction horizon to the same extent.

\begin{table}[t]
    \centering
    \small
    \begin{tabular}{c|ccc|ccc|ccc}
           \multicolumn{1}{c}{} & \multicolumn{9}{c}{Prediction horizon} \\ 
          Prediction & \multicolumn{3}{c|}{15 min} & \multicolumn{3}{c|}{30 min} & \multicolumn{3}{c}{45 min}   \\ \cline{2-10}
          interval & ICP & MIL & RMSE & ICP & MIL & RMSE & ICP & MIL & RMSE \\ \hline \hline
95\% & 94\% & 591 & \multirow{5}{*}{152.75}
     & 92\% & 591 & \multirow{5}{*}{164.58}
     & 92\% & 591 & \multirow{5}{*}{168.91} \\ \cline{1-3} \cline{5-6} \cline{8-9}
90\% & 89\% & 498 & 
     & 86\% & 498 & 
     & 86\% & 498 &  \\ \cline{1-3} \cline{5-6} \cline{8-9}
80\% & 79\% & 389 & 
     & 77\% & 388 & 
     & 77\% & 388 &  \\ \cline{1-3} \cline{5-6} \cline{8-9}
60\% & 64\% & 256 & 
     & 61\% & 255 & 
     & 58\% & 255 &  \\ \cline{1-3} \cline{5-6} \cline{8-9}
20\% & 29\% & 77 & 
     & 25\% & 77 & 
     & 21\% & 77 &  \\ \hline
    \end{tabular}
    \caption{DQR results for vehicle run time across all 39 links of 300S. MIL and RMSE in seconds.}
    \label{tab:DQR_results}
\end{table}

\begin{table}[t]
    \centering
    \small
    \begin{tabular}{c|ccc|ccc|ccc}
           \multicolumn{1}{c}{} & \multicolumn{9}{c}{Prediction horizon} \\ 
          Prediction & \multicolumn{3}{c|}{15 min} & \multicolumn{3}{c|}{30 min} & \multicolumn{3}{c}{45 min}   \\ \cline{2-10}
          interval & ICP & MIL & RMSE & ICP & MIL & RMSE & ICP & MIL & RMSE \\ \hline
95\% & 94\% & 653 & \multirow{5}{*}{179.43}
     & 75\% & 395 & \multirow{5}{*}{179.20}
     & 74\% & 381 & \multirow{5}{*}{178.95} \\ \cline{1-3} \cline{5-6} \cline{8-9}
90\% & 86\% & 493 & 
     & 65\% & 303 & 
     & 64\% & 295 &  \\ \cline{1-3} \cline{5-6} \cline{8-9}
80\% & 70\% & 341 & 
     & 51\% & 220 & 
     & 50\% & 215 &  \\ \cline{1-3} \cline{5-6} \cline{8-9}
60\% & 49\% & 200 & 
     & 37\% & 138 & 
     & 37\% & 135 &  \\ \cline{1-3} \cline{5-6} \cline{8-9}
20\% & 20\% & 56 & 
     & 17\% & 40 & 
     & 17\% & 40 &  \\ \hline
    \end{tabular}
    \caption{BRNN results for vehicle run time across all 39 links of 300S. MIL and RMSE in seconds.}
    \label{tab:brnn_results}
\end{table}

\begin{table}[t]
    \centering
    \small
    \begin{tabular}{c|ccc|ccc|ccc}
           \multicolumn{1}{c}{} & \multicolumn{9}{c}{Prediction horizon} \\ 
          Prediction & \multicolumn{3}{c|}{15 min} & \multicolumn{3}{c|}{30 min} & \multicolumn{3}{c}{45 min}   \\ \cline{2-10}
          interval & ICP & MIL & RMSE & ICP & MIL & RMSE & ICP & MIL & RMSE \\ \hline
95\% & 81\% & 412 & \multirow{5}{*}{156.92}
     & 87\% & 525 & \multirow{5}{*}{177.47}
     & 89\% & 619 & \multirow{5}{*}{182.70} \\ \cline{1-3} \cline{5-6} \cline{8-9}
90\% & 75\% & 347 & 
     & 80\% & 442 & 
     & 84\% & 521 &  \\ \cline{1-3} \cline{5-6} \cline{8-9}
80\% & 65\% & 271 & 
     & 70\% & 345 & 
     & 75\% & 407 &  \\ \cline{1-3} \cline{5-6} \cline{8-9}
60\% & 49\% & 178 & 
     & 55\% & 227 & 
     & 59\% & 267 &  \\ \cline{1-3} \cline{5-6} \cline{8-9}
20\% & 23\% & 54 & 
     & 25\% & 68 & 
     & 26\% & 80 &  \\ \hline
    \end{tabular}
    \caption{Kalman filter results for vehicle run time across all 39 links of 300S. MIL and RMSE in seconds.}
    \label{tab:kalman_results}
\end{table}

When evaluating the ICP metrics of the multi-link travel time uncertainty, we see that the DQR model performs very well for the 80\%, 90\% and 95\% prediction intervals, both for the 15 minute time horizon ($t + 1$), but also for the 30 and 45 minutes time horizons ($t + 2$ and $t + 3$), with a constant, but very small underestimation of the uncertainty interval (1-4 pp.). For the 20\% and 60\%, we see more inconsistent results for the DQR model, with both over- and underestimation of the uncertainty interval. The BRNN model only consistently outperforms the DQR wrt. ICP across the prediction time horizon for the 20\% prediction interval, while also predicts reasonable for the 15 minute horizon ($t + 1$) for all prediction intervals. The BRNN is though clearly better than the linear Kalman filter when evaluating the ICP metric at all prediction intervals for the 15 minute horizon ($t + 1$), while the performance results for the BRNN at 30 minutes and 45 minutes time horizons ($t + 2$ and $t + 3$) are worse than the linear Kalman filter. This again supports the conclusion that the BRNN model is not able to learn the time-dependent uncertainty with respect to the prediction horizon.

\subsubsection{Decision support system for connection assurance}
\label{sec:decision_support_system}

\begin{figure}[t]
    \centering
    \includegraphics[width=11cm]{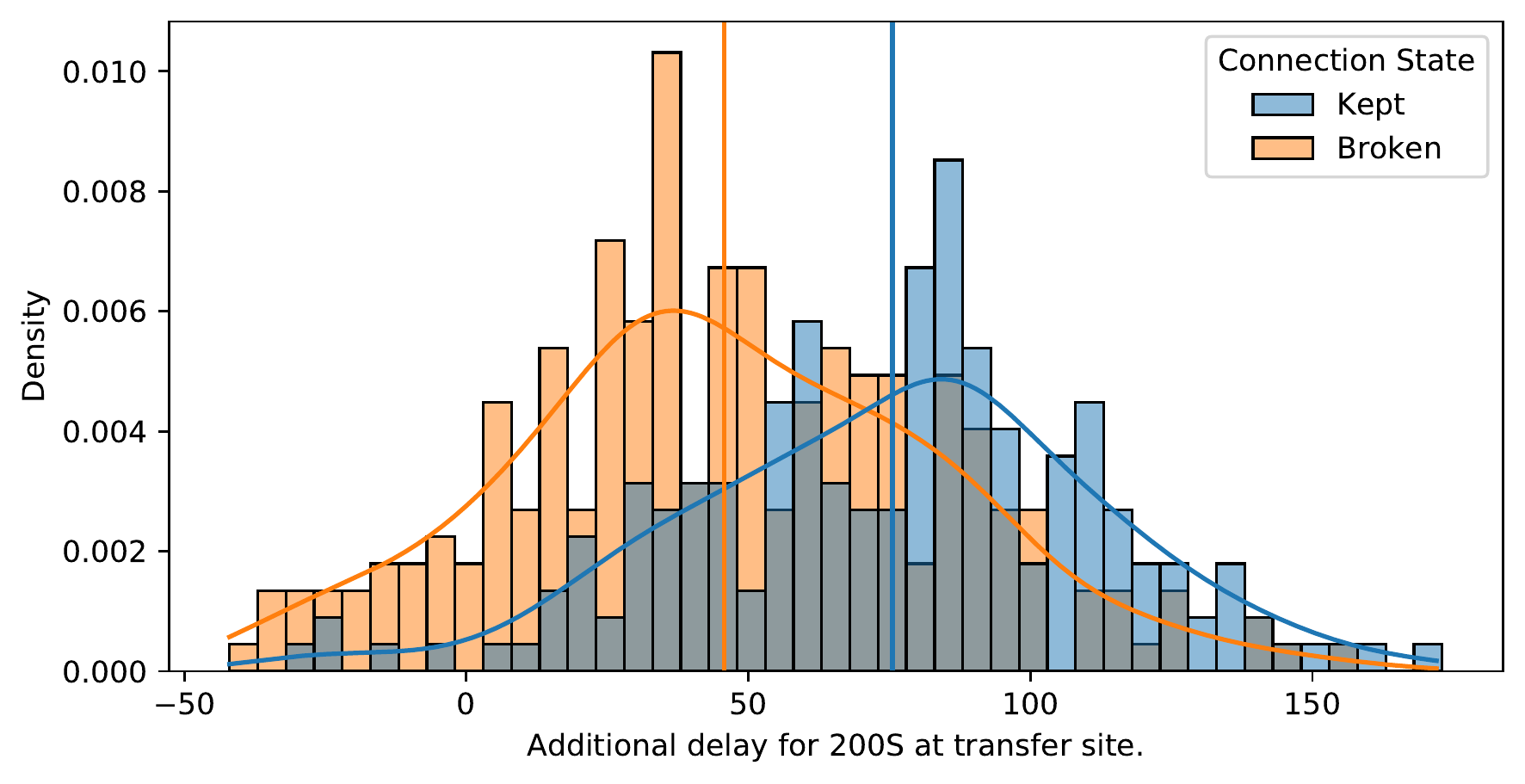}
    \caption{Introduced additional delay for keeping or breaking the connection at the use case transfer site.}
    \label{fig:additional-delay-200}
\end{figure}

To finalize our experiment, we demonstrate the usefulness of the presented models as part of a decision support system for connection assurance with intelligent uncertainty handling. We match vehicle journeys on each line by the scheduled connection as illustrated in \Cref{fig:connection_assurance} (left). In total, 446 journeys are matched from our test dataset, having a scheduled connection at the use case transfer site. \Cref{fig:additional-delay-200} shows the additional delay introduced on the 200S line when the connection is either kept or broken. This is fully aligned with \Cref{fig:connection_assurance} (center and right, respectfully), where keeping the connection can result in additional delays for either lines. Specifically, we see a mean additional delay induced on the transfer site of 48 seconds when the connection is broken, and 81 seconds when it is kept, i.e.\ an average additional delay of 69\% for keeping the connection. Most of this time could just as well be used at the origin of the 200S line, only a few stops upstream, thus allowing more people to board the line there, which is key for passenger satisfaction given that this is a urban train station. 
We therefore aim at helping the driver of the 200S line to decide when to depart from the train station to minimize any additional delay introduced at the transfer site, while keeping the connection. This will allow more passengers to transfer from the train to the bus, and not delay any passenger travel time - i.e.\ a holding strategy that prioritizes holding at the train station over holding at the transfer site. 

\begin{figure}[t]
    \centering
    \includegraphics[width=\textwidth]{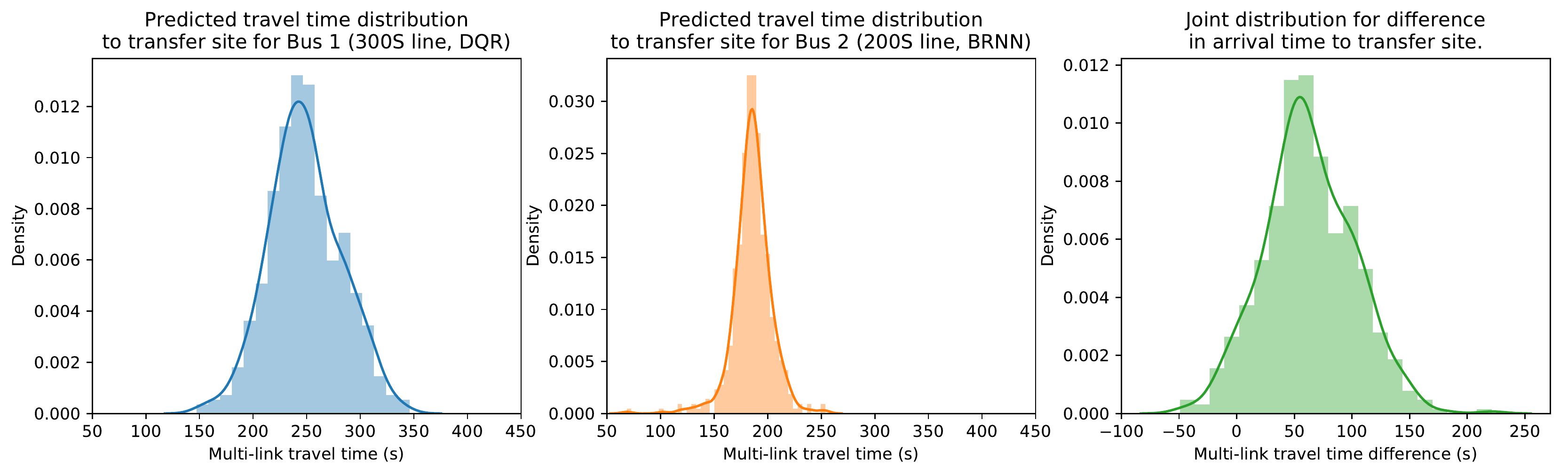}
    \caption{Example of predicted multi-link travel time distributions for each vehicle included in a scheduled connection, their difference which can be used as a decision variable for bus holding strategy.}
    \label{fig:joint-arrival}
\end{figure}

This decision is to be based on both the travel time uncertainty of the 200S line, between the train station (current position) and the transfer site, and the travel time uncertainty of the 300S line from its current position to the transfer site. \Cref{fig:joint-arrival} shows the predicted multi-link travel time distribution for each vehicle for one of the the 446 scheduled connections, and the difference cf. \cref{eq:connection_prop}. Notice that, here, we can mix the two different proposed link prediction models. Specifically, in this case, we achieved the best results by using the DQR model for the 300S line, and the BRNN model for the 200S line. The difference in travel time is exactly a distribution over time that can be prioritized for this control strategy, where 0 indicates the probability of both vehicles arriving perfectly synchronous at the transfer site. In our experiment, we prioritize this time with 80\% for holding at the train station and collecting additional passengers, and 20\% for holding at the transfer site. This split is arbitrarily chosen to demonstrate the advantages of \textit{uncertainty-aware} handling of connection assurance. In real-world applications, the prioritization would be a result of many factors, including passenger demand, how many buses a single stop point can hold, etc.

\begin{figure}[t]
    \centering
    \includegraphics[width=11cm]{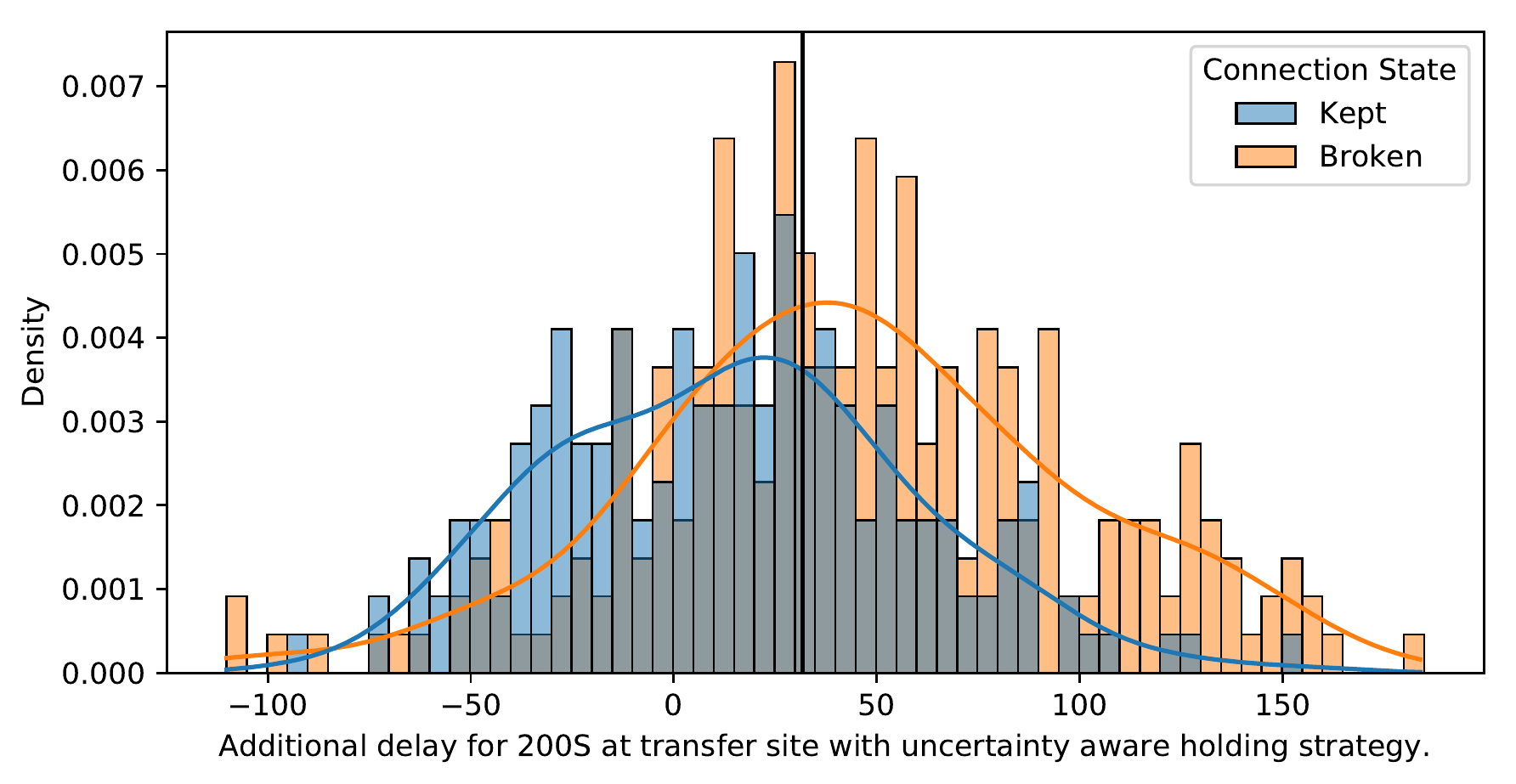}
    \caption{Introduced additional delay at the use case transfer site with \textit{uncertainty-aware} handling of connection assurance.}
    \label{fig:additional-delay-200-with-holding}
\end{figure}

The result of imposing the described decision support system onto the test dataset is shown in \Cref{fig:additional-delay-200-with-holding}. Note that the distinction by connection state still reflects the original data, and it is simply retained for comparison to \Cref{fig:additional-delay-200}. The main finding is that, by prioritizing the travel time uncertainty, we have in general reduced the additional delay introduced at the transfer site. We still see a positive introduced mean delay, i.e.\ the 200S will still more often wait for the 300S line, than vice-versa. This is expected since the system prioritized 20\% of the time uncertainty for this task, but the mean delay introduced with the proposed uncertainty aware holding strategy is reduced to 32 seconds, which is well below both the mean for the previously kept and broken connections, 81 seconds, and 48 seconds respectively, corresponding to a reduction of $60$\% and $33\%$. 

\section{Conclusion}
\label{sec:conclusion}
This paper has presented two approaches for uncertainty estimation adapted and extended for the bus travel time problem: \textit{Deep quantile regression} (DQR) and \textit{Bayesian recurrent neural networks} (BRNN). The output for both models are thus accurate uncertainty estimates for the time needed for a vehicle to travel from its current position to a specific downstream stop point or transfer site.

Our results for evaluating the multi-link travel time uncertainty showed that the DQR model performs very well for the 80\%, 90\% and 95\% prediction intervals, both for the 15 minute time horizon ($t + 1$), but also for the 30 and 45 minutes time horizon ($t + 2$ and $t + 3$), with a constant, but very small underestimation of the uncertainty interval (1-4 pp.). However, we also showed examples of how the BRNN model was able to capture the uncertainty of some specific cases better than the DQR model. Overall, we observed significant better performance for point prediction from the DQR model, both compared to the BRNN and the Kalman filter.

Furthermore, to motivate and support our arguments for the modelling of travel time uncertainty, we have presented an \textit{uncertainty-aware} decision support system for public transport connection assurance, specifically by intelligently prioritizing the uncertainty of travel time at different strategic holding points. Based on data from Copenhagen, our experiment showed that we could reduce the expected additional introduced delay at a transfer site from 48-81 seconds to 32 seconds.

Besides the design of these neural networks, our main contributions are: i) the comparison between DQR and BRNN performance for short-term public transport travel time uncertainty estimation, ii) the presented sampling technique in order to aggregate quantile estimates from DQR model for link-level travel time to yield the \textit{multi-link travel} time distributions, and iii) the demonstration of using uncertainty-aware models for transfer synchronization and connection assurance.

Our conclusion is that both DQR and BRNN are applicable in the real-time prediction systems used by many public transport operators around the world. The training of these models is obviously more computational expensive, than the point-estimate models currently used, but in the case of DQR the additional training time is negligible. At run-time they both add very little computational complexity. Since public transport, and in special bus transport, is prone to variability in travel time, it is hard to argue, why such systems should not model and use the uncertainty. 

\subsection{Future work}
Our work makes use of 1D-spatial convolutions, one interesting extension to this research would be to model the bus-network as a graph neural network, allowing the spatial correlations to propagate in the bus network. We would also like actual drivers to test out proposed decision support system. Currently, the presented results are based on the assumption that the driver proceeds exactly as the system proposes, but is often not the case in real-world scenarios. It would be interesting to show how sensitive the results are to the drivers adherence to the support system, i.e.\ how large part of the effect we have demonstrated in this work is measurable.

Finally, we would like to extend the \textit{uncertainty-aware} connection assurance decision system, taking into account a public transport route might have multiple scheduled transfer synchronizations and sites.

\section*{Funding sources and declaration of interests}
This research did not receive any specific grant from funding agencies in the public, commercial, or not-for-profit sectors. \\

Declarations of interest: none

\bibliography{library,local}
\newpage

\end{document}